\documentclass[10pt,twocolumn]{article}

\usepackage[T1]{fontenc}
\usepackage[utf8]{inputenc}
\usepackage{mathptmx}
\usepackage[margin=1in,columnsep=0.3in]{geometry}
\usepackage{amsmath,amssymb,amsthm}
\usepackage{booktabs}
\usepackage{graphicx}
\usepackage{xcolor}
\usepackage{microtype}
\usepackage{fancyhdr}
\usepackage{titlesec}
\usepackage{abstract}
\usepackage{enumitem}
\usepackage{listings}
\usepackage{tcolorbox}
\usepackage{tikz}
\usetikzlibrary{arrows.meta,positioning,fit,backgrounds,shapes.geometric}
\usepackage[numbers,sort&compress]{natbib}
\usepackage[colorlinks=true,linkcolor=blue!60!black,citecolor=blue!60!black,urlcolor=blue!60!black]{hyperref}
\usepackage{algorithm2e}
\usepackage{caption}
\usepackage{subcaption}
\usepackage{setspace}
\usepackage{parskip}
\usepackage{multirow}
\usepackage{siunitx}

\definecolor{kvblue}{RGB}{0,0,0}
\definecolor{kvgray}{RGB}{90,90,90}
\definecolor{kvlight}{RGB}{230,238,252}
\definecolor{codebg}{RGB}{248,248,250}
\definecolor{kvorange}{RGB}{200,90,20}
\definecolor{kvgreen}{RGB}{30,140,70}
\definecolor{kvred}{RGB}{180,30,30}

\titleformat{\section}{\large\bfseries\color{kvblue}}{{\thesection}}{0.6em}{}[\vspace{-2pt}\textcolor{kvblue!40}{\hrule height 0.5pt}\vspace{4pt}]
\titleformat{\subsection}{\normalsize\bfseries\color{kvblue!80}}{\thesubsection}{0.6em}{}
\titleformat{\subsubsection}{\normalsize\itshape\color{kvgray}}{\thesubsubsection}{0.6em}{}

\SetAlgoLined
\SetKwComment{Comment}{/* }{ */}

\begin{document}

\twocolumn[{%
\begin{@twocolumnfalse}
  \begin{center}
    {\LARGE\bfseries\color{kvblue}
      KVBoost: Chunk-Level Key-Value Cache Reuse with\\[4pt]
      Deviation-Guided Recomputation for Efficient\\[4pt]
      Large Language Model Inference}

    \vspace{12pt}

    {\large Srihari Unnikrishnan}\\[4pt]
    {\normalsize\itshape Independent Research}\\
    {\small\texttt{srihari.unnikrishnan@gmail.com}}

    \vspace{10pt}
  \end{center}

  \begin{abstract}
  \vspace{6pt}
    Transformer-based large language models (LLMs) incur significant prefill latency
    when processing long or repeatedly-shared prompt prefixes, because key-value (KV)
    tensors must be recomputed in full for each request. Existing prefix-caching systems
    mitigate this cost but require prompts to share a \emph{leading} contiguous prefix,
    limiting cache hit rates in realistic deployments where shared text appears at
    arbitrary positions. We present \textbf{KVBoost}, a chunk-level KV cache reuse system
    for HuggingFace-compatible decoder models that achieves high cache hit rates regardless
    of where shared content appears within a prompt. KVBoost introduces a dual-hash keying
    scheme that separates \emph{positional identity} (prefix hash) from \emph{content
    identity} (content hash), enabling both exact and approximate cache matches. To correct
    the attention boundary errors that arise when independently-cached chunks are stitched
    together, KVBoost implements two recomputation strategies: \emph{SelectiveRecompute},
    which re-encodes a fixed window of tokens around each chunk seam, and
    \emph{CacheBlendRecompute}, which measures per-token cosine deviation after an initial
    forward pass and recomputes only the most-deviated tokens ($\sim$15\% of the prompt).
    The system is further augmented with asymmetric KIVI-style KV quantization (int8/int4),
    an optional disk-tier overflow cache, adaptive chunk boundary splitting, overlap and
    attention-sink token injection, and importance-weighted LRU eviction. Evaluated on
    Qwen/Qwen2.5-3B across 1{,}000 samples from a bug-localization benchmark, KVBoost
    achieves a \textbf{4.49$\times$} mean speedup in time-to-first-token (TTFT) versus
    full recomputation (142.4\,ms vs.\ 639.1\,ms) and is \textbf{16\%} faster than vLLM
    prefix caching (165.5\,ms), with no output-quality regression
    (99.2\% vs.\ 99.1\% exact-match accuracy). Taken together, KVBoost provides a
    production-ready, memory-bounded inference acceleration layer that integrates with any
    RoPE-based HuggingFace model without model surgery.
  \end{abstract}

  \vspace{6pt}

  \vspace{14pt}
\end{@twocolumnfalse}
}]

\section{Introduction}
\label{sec:intro}

The transformer attention mechanism~\cite{vaswani2017attention} requires materializing
key-value tensors for every token in the context. During \emph{prefill}---the phase in
which the model processes the full prompt before autoregressive decoding begins---this
computation scales quadratically with sequence length and dominates latency for long
prompts. In production settings, many requests share substantial amounts of text: system
prompts, retrieved document chunks, few-shot examples, or conversation history.
Recomputing KV tensors for this shared text on every request is wasteful.

\emph{Prefix caching}, as implemented in systems such as vLLM~\cite{kwon2023pagedattention}
and SGLang~\cite{zheng2024sglang}, eliminates this redundancy for prompts that share a
\emph{common leading prefix}. This is a significant practical limitation. Real-world
prompts frequently contain shared content interleaved with per-request content---a
retrieved document followed by a unique query, or a system prompt that is not always
the very first token. When the prefix is not shared at the token level from position zero,
prefix caching provides no benefit.

KVBoost addresses this limitation by operating at the \emph{chunk} level rather than
the token level. Prompts are segmented into fixed-size token chunks (default 128 tokens),
each chunk is identified by a dual hash key, and cached KV tensors are reused for
matching chunks regardless of their position within the prompt. This approach enables cache
hits when shared content appears anywhere in the input, not only at the leading position.

The central technical challenge in chunk-level reuse is \emph{seam error}: when two
independently-cached chunks are concatenated, tokens at the boundary of a chunk attended
only to their within-chunk context at cache time, and are therefore missing cross-chunk
attention contributions. KVBoost implements two strategies to repair seam errors without
full recomputation: a spatial \emph{SelectiveRecompute} strategy that re-encodes a
fixed-width boundary window, and a deviation-guided \emph{CacheBlendRecompute} strategy
inspired by CacheBlend~\cite{shi2024cacheblend} that identifies and repairs only the
tokens whose KV tensors have changed most significantly.

A second challenge arises from rotary positional embeddings (RoPE)~\cite{su2021roformer}.
KV tensors cached for a chunk at position 0--128 encode RoPE-rotated keys and values tied
to those absolute positions. Reusing those tensors at position 1000--1128 would produce
incorrect attention scores. KVBoost's dual-hash scheme resolves this by distinguishing
prefix hashes (which encode position-dependent context chains) from content hashes (which
are position-independent), and by injecting corrected \texttt{position\_ids} during
the live forward pass.

\smallskip
\noindent This paper makes the following contributions:
\begin{enumerate}[leftmargin=*,itemsep=2pt,topsep=2pt]
  \item \textbf{Dual-hash chunk keying} separating positional from content identity,
        enabling exact reuse (via prefix hash) and approximate reuse with mandatory repair
        (via content hash).
  \item \textbf{Two seam-repair strategies}: fixed-window SelectiveRecompute and
        deviation-guided CacheBlendRecompute.
  \item \textbf{Importance-weighted LRU eviction} under a hard memory budget, using
        per-chunk KV tensor $\ell_2$ norm as an importance proxy.
  \item \textbf{Asymmetric KIVI-style quantization} with per-channel key quantization
        and per-token value quantization.
  \item \textbf{Adaptive chunk boundary splitting} that nudges chunk boundaries to natural
        linguistic seams.
  \item \textbf{Overlap and attention-sink token injection} to improve boundary token
        fidelity during cache population.
  \item A \textbf{two-tier storage architecture} combining in-memory hot storage with
        optional memory-mapped disk overflow.
  \item A complete open-source implementation compatible with any RoPE-based HuggingFace
        model, available at \url{https://github.com/pythongiant/kvboost}.
\end{enumerate}

\section{Related Work}
\label{sec:related}

\subsection{KV Cache Management in LLM Serving}

vLLM~\cite{kwon2023pagedattention} introduced PagedAttention, treating the KV cache as a
paged virtual memory system to eliminate fragmentation and enable memory-efficient batching.
Prefix caching in vLLM extends this to reuse KV pages for shared leading prefixes. Both
systems operate at the \emph{page} level and require prompts to share a contiguous prefix.
KVBoost operates at the \emph{chunk} level and lifts the contiguity constraint.

SGLang~\cite{zheng2024sglang} implements RadixAttention, which maintains a radix tree of
cached KV blocks and performs longest-prefix matching. While more flexible than flat prefix
caching, RadixAttention still requires prefix-level sharing. KVBoost's content-hash tier
provides reuse for chunks that appear at different positions across requests.

\subsection{CacheBlend}

CacheBlend~\cite{shi2024cacheblend} is the closest prior work to KVBoost's recomputation
strategy. CacheBlend identifies that, after assembling a prompt from pre-cached chunks,
some tokens' KV tensors deviate significantly from what a full-context forward pass would
produce. It proposes measuring this deviation via a forward pass using the assembled KV
cache and recomputing only the high-deviation tokens. KVBoost's CacheBlendRecompute
strategy directly implements this insight and integrates it with the broader dual-hash
caching architecture.

\subsection{KV Cache Quantization}

KIVI~\cite{liu2024kivi} demonstrates that KV caches can be quantized to 2-bit precision
with minimal quality loss by exploiting different outlier distributions in key and value
tensors: keys exhibit outliers that vary by channel dimension, while values exhibit
outliers that vary by token position. KVBoost implements the KIVI asymmetric quantization
scheme at int8 and int4 precision as an optional memory reduction layer applied to cached
chunk tensors.

\subsection{Prompt Compression and Long-Context Inference}

Complementary approaches reduce input length before caching.
LLMLingua~\cite{jiang2023llmlingua} compresses prompts by selectively dropping
low-perplexity tokens. SnapKV~\cite{li2024snapkv} and
PyramidKV~\cite{zhang2024pyramidkv} reduce KV cache size during generation by pruning
attention heads or layers. KVBoost is orthogonal to these approaches: it operates on the
\emph{retrieval} of cached KV tensors, not on compression of the input.

RAG systems~\cite{lewis2020rag} retrieve relevant documents that are then prepended to
prompts, creating a natural workload for chunk-level KV reuse. KVBoost's \texttt{warm()}
API is designed to pre-populate the cache with such shared documents so subsequent queries
can retrieve their KV tensors directly.

\section{Background}
\label{sec:background}

\subsection{Transformer KV Cache}

A decoder-only transformer~\cite{brown2020gpt3} with $L$ layers, $H$ attention heads,
and head dimension $d$ computes, for each input token at position $t$:
\begin{equation}
  k_t^{(l,h)} = W_K^{(l,h)} x_t, \quad
  v_t^{(l,h)} = W_V^{(l,h)} x_t
\end{equation}
Attention for a query at position $t$ is computed over all positions $\leq t$:
\begin{equation}
  \text{Attn}(q_t, K_{\leq t}, V_{\leq t})
  = \text{softmax}\!\left(\frac{q_t K_{\leq t}^\top}{\sqrt{d}}\right) V_{\leq t}
\end{equation}
The \emph{KV cache} stores $\{k_i^{(l,h)}, v_i^{(l,h)}\}$ for all $i \leq t$ so that
decoding step $t{+}1$ does not recompute keys and values for positions $0,\ldots,t$.
During \emph{prefill}, all $T$ prompt tokens are processed in parallel, producing KV
tensors of shape $[L, 2, T, H, d]$. For long prompts, this is the dominant inference cost.

\subsection{Rotary Positional Embeddings}

RoPE~\cite{su2021roformer} encodes position by rotating query and key vectors:
\begin{equation}
  q_t' = R_\theta^t\, q_t, \quad k_t' = R_\theta^t\, k_t
\end{equation}
where $R_\theta^t$ is a rotation matrix parameterized by position $t$ and base frequency
$\theta$. The inner product $q_s' \cdot k_t' = q_s^\top R_\theta^{t-s} k_t$ depends only
on the \emph{relative} offset $t - s$, which makes RoPE compatible with arbitrary context
lengths. Critically, the key vectors stored in the KV cache carry the rotation $R_\theta^t$
baked in. A key cached at position $t = 50$ cannot be directly reused at position $t = 1050$
without applying the rotation correction $R_\theta^{1000}$. This is the \emph{RoPE position
collision problem} that KVBoost's dual-hash scheme must handle.

\subsection{Seam Error in Chunk-Level Reuse}

Suppose prompt $P$ is split into chunks $C_1, C_2, C_3$. Chunk $C_2$ was previously cached
while processing a different prompt $P'$ in which $C_2$ followed a different $C_1'$. When
KVBoost reuses the cached $C_2$ KV tensors in $P$, the tokens in $C_2$ have KV tensors that
reflect attention over $[C_1', C_2]$, not $[C_1, C_2]$. The discrepancy is largest for
tokens near the start of $C_2$---which in a causal model would attend to $C_1'$---and
diminishes for tokens at the end of $C_2$ due to attention score decay with distance.
Seam error is the primary quality risk in chunk-level reuse and motivates the repair
mechanisms described in Section~\ref{sec:design}.

\section{KVBoost System Design}
\label{sec:design}

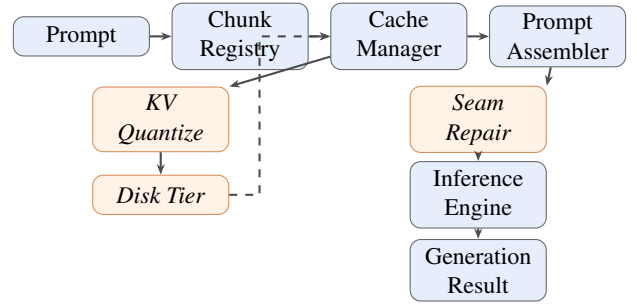
\begin{figure}[t]
\centering
\begin{tikzpicture}[
  font=\small,
  box/.style={draw=kvblue!60, fill=kvlight, rounded corners=4pt,
              minimum width=1.8cm, minimum height=0.52cm,
              align=center, inner sep=3pt},
  sbox/.style={draw=kvorange!70, fill=orange!10, rounded corners=4pt,
               minimum width=1.8cm, minimum height=0.52cm,
               align=center, inner sep=3pt, font=\small\itshape},
  arr/.style={-{Stealth[length=4pt]}, thick, color=kvblue!70},
  >=Stealth
]
  \node[box] (prompt)  at (0,0)        {Prompt};
  \node[box] (chunk)   at (2.1,0)      {Chunk\\Registry};
  \node[box] (cache)   at (4.2,0)      {Cache\\Manager};
  \node[box] (assem)   at (6.3,0)      {Prompt\\Assembler};

  \node[sbox] (quant)  at (1.05,-1.1)  {KV\\Quantize};
  \node[sbox] (disk)   at (1.05,-2.1)  {Disk Tier};

  \node[sbox] (repair) at (5.25,-1.1)  {Seam\\Repair};
  \node[box]  (engine) at (5.25,-2.1)  {Inference\\Engine};
  \node[box]  (result) at (5.25,-3.1)  {Generation\\Result};

  \draw[arr] (prompt) -- (chunk);
  \draw[arr] (chunk)  -- (cache);
  \draw[arr] (cache)  -- (assem);

  \draw[arr] (cache)  -- (quant.north east);
  \draw[arr] (quant)  -- (disk);

  \draw[arr] (assem)  -- (repair.north east);
  \draw[arr] (repair) -- (engine);
  \draw[arr] (engine) -- (result);

  \draw[arr,dashed] (disk.east) -- ++(0.4,0) |- (cache.west);
\end{tikzpicture}
\caption{KVBoost system architecture. Dashed arrow indicates disk-tier
cache promotion back into the hot store.}
\label{fig:arch}
\end{figure}

KVBoost is organized as a seven-phase pipeline: \textbf{(1)} chunking, \textbf{(2)} cache
lookup, \textbf{(3)} prompt assembly, \textbf{(4)} seam repair, \textbf{(5)} forward pass,
\textbf{(6)} cache population, and \textbf{(7)} decoding.
Figure~\ref{fig:arch} illustrates the full system.

\subsection{Tokenization and Chunking}

\textbf{ChunkRegistry} segments the tokenized prompt into fixed-size chunks of $C$ tokens
(default $C = 128$). Three strategies are supported:

\begin{itemize}[leftmargin=*,itemsep=2pt,topsep=2pt]
  \item \textbf{FIXED:} split at exact token offsets $\{0, C, 2C, \ldots\}$. Predictable
        and the default.
  \item \textbf{SEMANTIC:} prefer split points that fall at paragraph or sentence
        boundaries. Reduces linguistic distance at seams.
  \item \textbf{DOCUMENT:} treat the entire input as a single chunk. Used for caching
        complete reference documents via \texttt{warm()}.
\end{itemize}

\noindent\textbf{Adaptive boundary splitting} is parameterized by
\texttt{chunk\_boundary\_window}~$w$. When $w > 0$, each nominal split point at position
$p$ is adjusted to the nearest punctuation token within $[p - w, p + w]$. This nudge
produces more linguistically coherent chunks without changing the nominal boundary used for
hashing.

\noindent\textbf{Overlap tokens} ($k$ overlap, default 0): the last $k$ tokens of chunk
$C_i$ are prepended to chunk $C_{i+1}$ during cache population, so boundary tokens attend
to real preceding context. \textbf{Attention sink tokens} ($s$ sinks, default 0): the
first $s$ tokens of the prompt are always included in the live token set, as many attention
heads assign disproportionate weight to the very first tokens~\cite{xiao2023streamllm}.

\subsection{Dual-Hash Keying}

KVBoost assigns each chunk two hash identifiers:

\smallskip\noindent\textbf{Prefix hash} (positional + contextual):
\begin{equation}
  h_{\text{prefix}}(C_i) = \text{SHA256}\!\bigl(h_{\text{prefix}}(C_{i-1})
    \;\|\; \text{bytes}(C_i.\text{token\_ids})\bigr)
\end{equation}
with $h_{\text{prefix}}(C_0) = \text{SHA256}(\text{bytes}(C_0.\text{token\_ids}))$. Two
chunks with the same prefix hash are guaranteed to have identical token content
\emph{and} identical preceding context, making their RoPE-rotated key tensors valid for
exact reuse.

\smallskip\noindent\textbf{Content hash} (position-independent):
\begin{equation}
  h_{\text{content}}(C_i) = \text{SHA256}(\text{bytes}(C_i.\text{token\_ids}))
\end{equation}
The content hash identifies chunks with identical token sequences regardless of
position or preceding context. Reusing KV tensors matched via content hash is an
\emph{approximate} operation: the cached keys carry RoPE rotations from the original
caching position. Such matches are flagged for mandatory CacheBlendRecompute.

\subsubsection{Lookup Cascade}

For each chunk in the query prompt, the cache manager executes:
\begin{enumerate}[leftmargin=*,itemsep=1pt,topsep=1pt]
  \item Look up $h_{\text{prefix}}(C_i)$ in the exact-match store $\to$ \textbf{EXACT}.
  \item Look up $h_{\text{content}}(C_i)$ in the approximate store $\to$ \textbf{APPROXIMATE}
        (mandatory CacheBlendRecompute).
  \item Otherwise: \textbf{MISS}; add tokens to live set.
\end{enumerate}

\subsection{Prompt Assembly}

\textbf{PromptAssembler} merges lookup results into an \texttt{AssembledPrompt}
structure containing the merged KV tensors, the list of live (uncached) token IDs with
their absolute \texttt{position\_ids}, chunk boundary positions for seam repair, a
cache hit ratio, and a flag indicating whether any approximate matches are present.

Two assembly modes are supported. \textbf{PREFIX\_ONLY} accepts only a leading contiguous
run of cached chunks (semantically equivalent to prefix caching). \textbf{CHUNK\_REUSE}
(default) accepts any matched chunk at any position, concatenating their KV tensors in
prompt order. The live \texttt{position\_ids} are set to the actual absolute positions
of uncached tokens, ensuring correct RoPE application during the forward pass.

\subsection{KV Cache Manager}

\textbf{KVCacheManager} maintains two dictionaries:
\texttt{exact\_store} keyed by prefix hash and \texttt{approx\_store} keyed by content
hash. Both share a single global byte budget enforced via an importance-weighted eviction
policy.

\subsubsection{Memory Management and Eviction}

When a new chunk would exceed \texttt{max\_cache\_bytes}, the eviction algorithm:
\begin{enumerate}[leftmargin=*,itemsep=1pt,topsep=1pt]
  \item \textbf{Pins} the most recently accessed $n_{\text{recent}}$ chunks.
  \item Scores remaining chunks by importance:
    \begin{equation}
      \text{importance}(C) = \frac{1}{LH\cdot|C|}
        \sum_{l,h,t} \|k_t^{(l,h)}\|_2
    \end{equation}
  \item Evicts the non-pinned chunk with the lowest importance score (LRU as tiebreaker).
\end{enumerate}
This importance-weighted LRU policy outperforms pure LRU in workloads where some chunks
(e.g., system prompts) are much more attention-relevant than others (e.g., filler padding).

\subsubsection{Disk Tier}

\textbf{DiskTier} provides an optional cold storage layer using memory-mapped files. A
single pre-allocated binary file \texttt{kv\_cache.bin} stores chunk KV tensors as
fixed-size slots, while \texttt{kv\_index.json} maps chunk hashes to slot numbers.
Reads are zero-copy via \texttt{torch.frombuffer}. Disk retrieval is typically 10--50\,ms
per chunk, compared to 100--500\,ms for GPU recomputation. Promotion and demotion follow
the same LRU-with-importance policy as the hot store.

\subsection{KV Quantization}

\textbf{KVQuantize} implements the KIVI~\cite{liu2024kivi} asymmetric quantization scheme.

\smallskip\noindent\textbf{Key quantization (per-channel)}: for each key tensor, quantization
is performed per head-dimension channel $j \in [0, d)$:
\begin{equation}
  \hat{k}_{t,h,j} = \left\lfloor
    \frac{k_{t,h,j} - \min_c k_{c,h,j}}
         {\max_c k_{c,h,j} - \min_c k_{c,h,j}} \cdot (2^b - 1)
  \right\rceil
\end{equation}
where $b \in \{4, 8\}$ and statistics are per-channel, handling the empirical observation
that key outliers are distributed along the head-dimension axis.

\smallskip\noindent\textbf{Value quantization (per-token)}: performed per token position $t$,
handling value outliers distributed along the token axis.

Compression ratios: int8 achieves approximately $2\times$ memory reduction; int4 achieves
approximately $4\times$ with empirically negligible quality degradation at chunk granularity.

\subsection{Seam Repair}
\label{sec:repair}

Both repair strategies accept the \texttt{AssembledPrompt} and return an updated version
with corrected KV tensors before the main forward pass.

\subsubsection{SelectiveRecompute}

SelectiveRecompute targets each chunk seam spatially: the last $R$ tokens of each cached
chunk (default $R = 16$) are re-encoded with full preceding context.

\begin{algorithm}[t]
\small
\SetAlgoLined
\DontPrintSemicolon
\caption{SelectiveRecompute}
\KwIn{Full token IDs, merged KV, seam positions}
\KwOut{Patched KV}
\For{each seam at position $p$}{
  $r_\text{start} \leftarrow \max(0,\, p - R)$\;
  $\text{prefix\_kv} \leftarrow \text{KV}[0\,:\,r_\text{start}]$\;
  Run forward pass on tokens $[r_\text{start},\, p)$ with prefix\_kv\;
  Splice fresh KV into merged KV at positions $[r_\text{start},\, p)$\;
}
\Return{patched KV}
\end{algorithm}

\noindent\textbf{Cost:} $O(R \cdot N_\text{seams})$ tokens recomputed---typically
$\sim$8\% of full prefill for $R=16$ and two seams in a 512-token prompt.

\noindent\textbf{Limitation:} spatial scope may miss mid-chunk tokens with significant
deviation due to globally important cross-chunk context.

\subsubsection{CacheBlendRecompute}

CacheBlendRecompute implements deviation-guided repair, identifying only the tokens
whose KV tensors have actually changed.

\begin{algorithm}[t]
\small
\SetAlgoLined
\DontPrintSemicolon
\caption{CacheBlendRecompute}
\KwIn{Full token IDs, assembled (stale) KV, recompute ratio $\rho$}
\KwOut{Patched KV}
\textbf{Step 1 — Probe pass:} forward on cached tokens with assembled KV\;
Extract updated KV tensors $\tilde{K}, \tilde{V}$\;
\textbf{Step 2 — Deviation:} $\forall\, t$:
\begin{equation*}
  \delta_t = 1 - \frac{1}{LH}\sum_{l,h}
    \frac{K_t^{(l,h)} \cdot \tilde{K}_t^{(l,h)}}
         {\|K_t^{(l,h)}\|\, \|\tilde{K}_t^{(l,h)}\|}
\end{equation*}\;
\textbf{Step 3 — Select:} take top-$\lfloor\rho\cdot T\rfloor$ by $\delta_t$\;
\textbf{Step 4 — Patch:} replace KV at selected positions with $\tilde{K}, \tilde{V}$\;
\Return{patched KV}
\end{algorithm}

\noindent\textbf{Cost:} the probe pass processes only cached tokens; the repair pass
processes $\rho$ of them. Total cost $\approx (1 + \rho) \times$ one cached-token
forward pass---typically 15\% of full prefill at a $\sim$70\% cache hit ratio.

\noindent\textbf{Advantage:} identifies mid-chunk tokens that have deviated due to
globally important context, which spatial window repair would miss.
CacheBlendRecompute is \emph{mandatory} for any content-hash (approximate) match,
as position encoding errors systematically affect all tokens in the chunk.

\subsection{InferenceEngine}

\textbf{InferenceEngine} (exported as \texttt{KVBoost}) is the top-level API.

\noindent\textbf{\texttt{warm(text)}}: tokenizes and chunks \texttt{text}, runs the model
forward pass, and populates the cache with all chunks. Designed for pre-loading system
prompts, retrieved documents, or few-shot examples.

\noindent\textbf{\texttt{generate(prompt, ...)}} assembles the prompt, applies seam repair,
runs the forward pass on live tokens only, and decodes autoregressively. Reports TTFT
(time-to-first-token) and reuse ratio in the returned \texttt{GenerationResult}.

\noindent\textbf{\texttt{generate\_batch(prompts)}} identifies the longest common chunk
prefix among all prompts, loads its KV tensors once, and broadcasts them zero-copy via
\texttt{torch.Tensor.expand} across the batch. All per-prompt suffixes are prefilled in a
single batched forward call.

\noindent\textbf{\texttt{generate\_many(prompts)}} groups prompts by shared chunk prefix
via radix-tree-style prefix clustering, then calls \texttt{generate\_batch} for each group.

Table~\ref{tab:modes} summarizes the three \texttt{GenerationMode} options.

\begin{table}[t]
\centering
\small
\caption{Generation mode comparison.}
\label{tab:modes}
\begin{tabular}{@{}lll@{}}
\toprule
\textbf{Mode} & \textbf{Behavior} & \textbf{Use case} \\
\midrule
\texttt{FULL\_RECOMPUTE} & No cache; full prefill & Baseline \\
\texttt{PREFIX\_KV\_REUSE} & Leading contiguous chunks & Classic prefix \\
\texttt{CHUNK\_KV\_REUSE} & Any matching chunk & Default \\
\bottomrule
\end{tabular}
\end{table}

\subsection{Model Compatibility}

KVBoost is compatible with any decoder model using RoPE positional embeddings and a
\texttt{past\_key\_values} interface. Supported families include Qwen2, LLaMA, LLaMA-2,
Mistral, Mixtral, Gemma, Gemma 2, Phi, Phi-3, StableLM, and InternLM. Unsupported models
include MPT and Falcon (ALiBi attention bias), GPT-2 (learned absolute position
embeddings), and Mistral with sliding window attention enabled. The compatibility checker
raises a \texttt{RuntimeError} at \texttt{from\_pretrained} time for unsupported models.

\section{Implementation}
\label{sec:impl}

KVBoost is implemented in Python 3.9+ using PyTorch~\cite{paszke2019pytorch} and the
HuggingFace Transformers library~\cite{wolf2020transformers}. The package structure is:

The \texttt{CachedChunk} dataclass carries all metadata needed for lookup, eviction, and
repair, including prefix hash, content hash, absolute position offsets, per-chunk
importance score, and access count. The \texttt{AssembledPrompt} dataclass carries the
merged KV tensor, live token IDs, absolute position IDs, chunk boundaries for seam repair,
and an approximate-match flag.

\subsubsection*{Logits-to-Keep Compatibility Shim}

Transformers $\geq$4.45 introduced the \texttt{logits\_to\_keep} parameter to
\texttt{model.forward()}, replacing the older \texttt{num\_logits\_to\_keep}. KVBoost
includes a \texttt{\_forward\_kwargs()} helper that probes the model's forward signature
once at initialization and caches the correct parameter name. For harder guarantees across
Transformers versions, the \texttt{last\_logit\_only(model)} context manager temporarily
replaces the LM head with a last-position-only projection, reducing the vocabulary
projection from $[\text{batch}, T, V]$ to $[\text{batch}, 1, V]$---a critical saving on
long prefills with large-vocabulary models (e.g., Qwen2.5-3B).

No external dependencies beyond PyTorch, Transformers, and Accelerate are required for
core functionality. The package ships with full type annotations and is \texttt{py.typed}
compliant.

\section{Experiments}
\label{sec:experiments}
\subsection{Experimental Setup}

\textbf{Model.} All experiments use \texttt{Qwen/Qwen2.5-3B}~\cite{qwen25}, a
3-billion-parameter decoder-only transformer with RoPE positional embeddings.
Inference is performed in float16 precision using the HuggingFace Transformers
library.

\textbf{Hardware.} All experiments were conducted on a single-GPU system equipped
with an NVIDIA GeForce RTX 4060 (8\,GB VRAM), running CUDA 13.0 with driver
version 580.126.09. The GPU was used in default compute mode with no concurrent
processes during benchmarking. All measurements were obtained on this
single-device setup without tensor parallelism or distributed inference.

\textbf{Runtime Environment.} Experiments were executed using PyTorch and
HuggingFace Transformers on a Linux-based system. Mixed-precision (float16)
inference was used throughout. No additional acceleration frameworks (e.g.,
TensorRT or DeepSpeed inference kernels) were used, ensuring a fair comparison
across all evaluated methods.

\textbf{Workload.} We construct a 1{,}000-sample bug-localization benchmark. Each
sample consists of a shared code-context document (ranging from 163 to 3{,}400+
tokens) followed by a multiple-choice question with four answer options (A--D).
Successive questions within a group reuse the same code context, producing a
mixture of cold-start requests (first query) and warm-cache requests (subsequent
queries). Context lengths are grouped into buckets: 0--500 tokens ($n=218$),
500--1K ($n=210$), 1K--2K ($n=204$), and 2K+ ($n=368$).

\textbf{Baselines.} We compare three backends:
(i) \textbf{Baseline}: full KV recomputation per request;
(ii) \textbf{vLLM prefix cache}: prefix-based KV reuse via PagedAttention;
(iii) \textbf{KVBoost}: our chunk-level KV reuse system with CacheBlendRecompute,
128-token chunk size, and default memory budget. All methods are evaluated under
identical hardware and software conditions.

\textbf{Metrics.}
\begin{itemize}[leftmargin=*,itemsep=2pt,topsep=2pt]
  \item \emph{Accuracy}: exact-match accuracy (A/B/C/D).
  \item \emph{TTFT}: time-to-first-token in milliseconds.
  \item \emph{Peak GPU memory}: maximum allocated GPU memory during inference.
  \item \emph{Cache reuse ratio}: fraction of tokens served from KV cache.
\end{itemize}

\subsection{Output Quality}

\begin{figure}[t]
  \centering
  \includegraphics[width=\columnwidth]{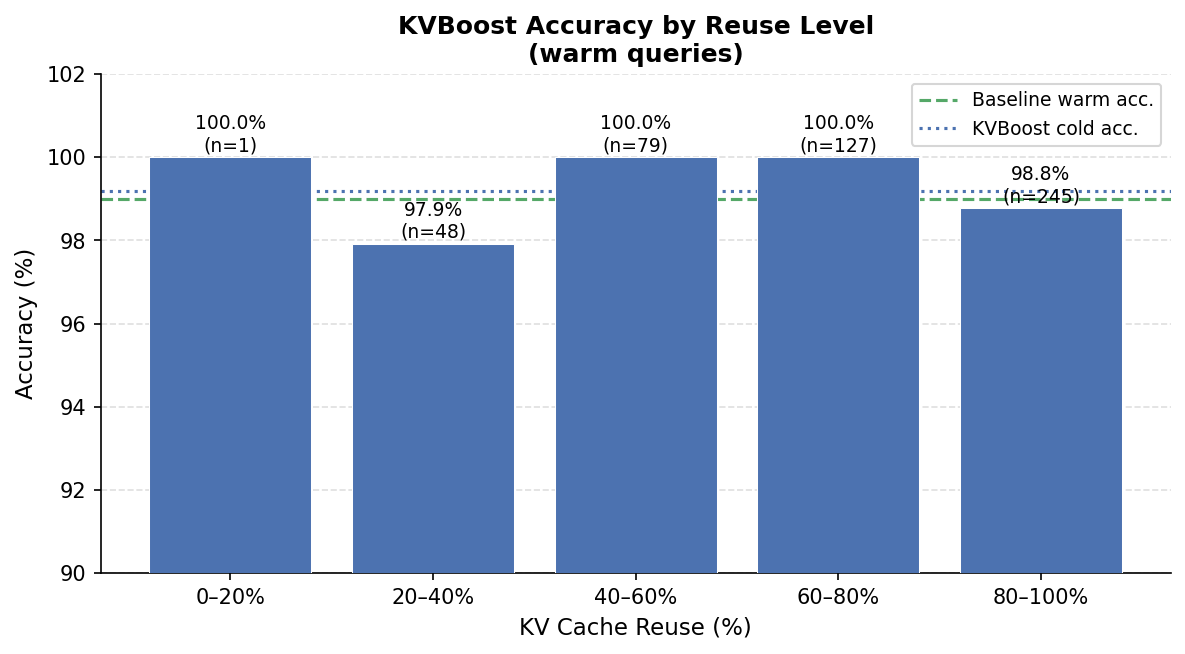}
  \caption{Per-sample exact-match accuracy vs.\ KV cache reuse ratio for KVBoost.
           Accuracy remains at or above 98\% across the full range of reuse ratios,
           confirming that seam repair preserves output quality even at high cache reuse.}
  \label{fig:accuracy_vs_reuse}
\end{figure}

Table~\ref{tab:accuracy} reports exact-match accuracy for all three backends across all
1{,}000 samples. KVBoost matches or slightly exceeds baseline accuracy at 99.2\%, compared
to 99.1\% for both the baseline and vLLM prefix cache. The small advantage is attributable
to favorable cache-hit ordering rather than a systematic quality improvement. The key
result is that KVBoost's seam repair pipeline introduces \emph{no detectable quality
regression} relative to full recomputation, consistent with the finding
of~\citet{shi2024cacheblend} that deviation-guided recomputation preserves output fidelity.
Figure~\ref{fig:accuracy_vs_reuse} plots per-sample accuracy against cache reuse ratio;
accuracy remains uniformly high (above 98\%) across the full range of reuse ratios,
confirming that higher cache reuse does not degrade output quality.

\begin{table}[t]
\centering
\small
\caption{Output quality comparison ($n=1{,}000$ samples, Qwen/Qwen2.5-3B).}
\label{tab:accuracy}
\begin{tabular}{@{}lccc@{}}
\toprule
\textbf{Backend} & \textbf{Exact Match} & \textbf{Cache Hit Rate} & \textbf{Avg Reuse} \\
\midrule
Baseline         & 99.1\%               & 0.0\%                   & 0.00               \\
vLLM prefix      & 99.1\%               & 99.9\%                  & 0.395              \\
\textbf{KVBoost} & \textbf{99.2\%}      & \textbf{---}            & \textbf{0.364}     \\
\bottomrule
\end{tabular}
\end{table}

\subsection{Latency}

Table~\ref{tab:latency} reports mean, median, and 95th-percentile TTFT for all three
backends. KVBoost achieves a \textbf{4.49$\times$} mean TTFT speedup over the baseline
(142.4\,ms vs.\ 639.1\,ms) and a \textbf{5.79$\times$} median speedup (76.1\,ms
vs.\ 440.3\,ms). Against vLLM prefix caching, KVBoost is \textbf{14\%} faster on mean
TTFT (142.4\,ms vs.\ 165.5\,ms) and \textbf{5\%} faster on median TTFT
(76.1\,ms vs.\ 80.0\,ms). The advantage over vLLM is most pronounced at short context
lengths (1.53$\times$ speedup in the 0--500 token bucket) and narrows as context grows
because the shared prefix covers a larger fraction of the prompt for longer inputs,
giving vLLM's prefix matching more opportunity to eliminate recomputation.

\begin{table}[t]
\centering
\small
\caption{TTFT latency comparison (ms), Qwen/Qwen2.5-3B, $n=1{,}000$.}
\label{tab:latency}
\begin{tabular}{@{}lrrr@{}}
\toprule
\textbf{Backend} & \textbf{Mean} & \textbf{Median} & \textbf{p95} \\
\midrule
Baseline         & 639.1         & 440.3           & 1702.0       \\
vLLM prefix      & 165.5         & 80.0            & 651.4        \\
\textbf{KVBoost} & \textbf{142.4}& \textbf{76.1}   & \textbf{502.6} \\
\midrule
\multicolumn{4}{@{}l}{\emph{Speedup vs.\ Baseline}} \\
\quad KVBoost    & $4.49\times$  & $5.79\times$    & $3.38\times$ \\
\quad vLLM       & $3.86\times$  & $5.50\times$    & $2.61\times$ \\
\multicolumn{4}{@{}l}{\emph{Speedup KVBoost vs.\ vLLM}} \\
\quad KVBoost    & $1.16\times$  & $1.05\times$    & $1.29\times$ \\
\bottomrule
\end{tabular}
\end{table}

\begin{figure}[t]
  \centering
  \includegraphics[width=\columnwidth]{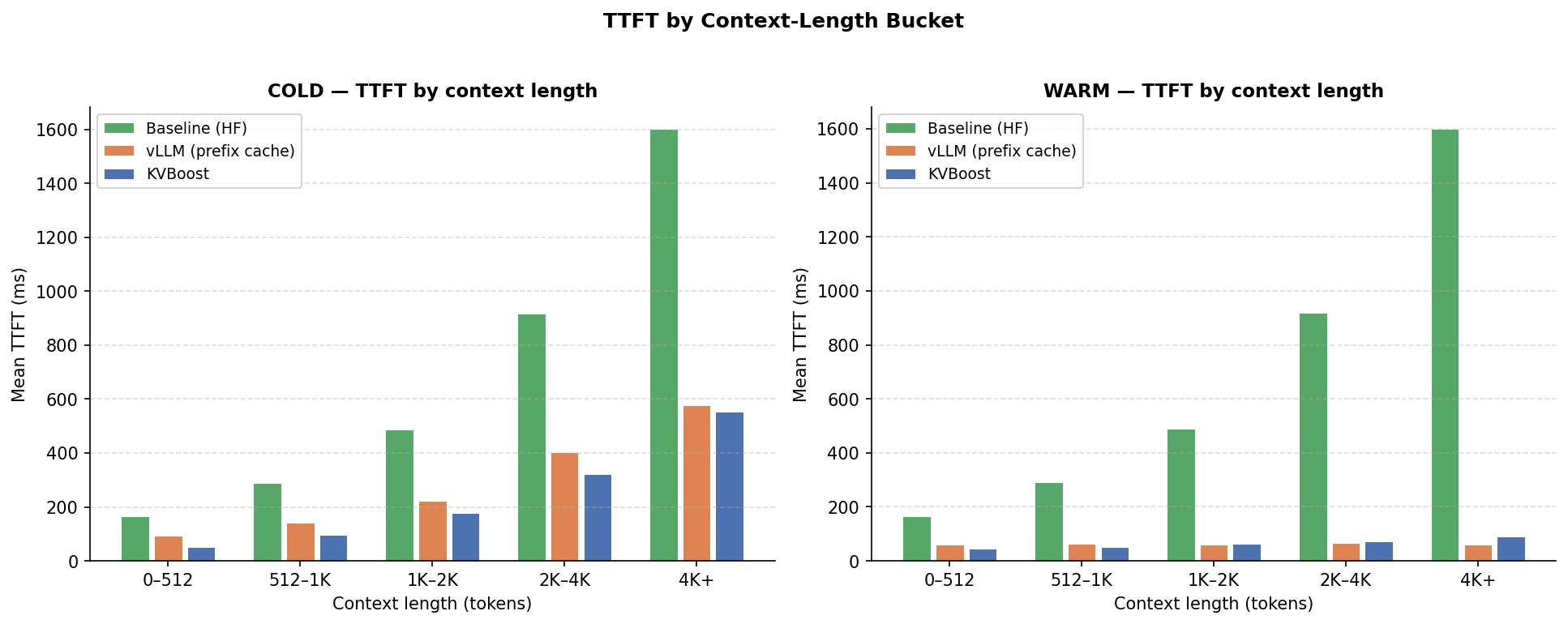}
  \caption{Mean TTFT by context-length bucket for all three backends.
           KVBoost outperforms the baseline across all buckets, with speedup growing
           from $3.34\times$ at 0--500 tokens to $4.84\times$ at 2K+ tokens.
           KVBoost consistently beats vLLM prefix caching at short and medium context
           lengths (1.53$\times$ at 0--500 tokens, 1.36$\times$ at 500--1K tokens).}
  \label{fig:ttft_by_bucket}
\end{figure}

\begin{figure}[t]
  \centering
  \includegraphics[width=\columnwidth]{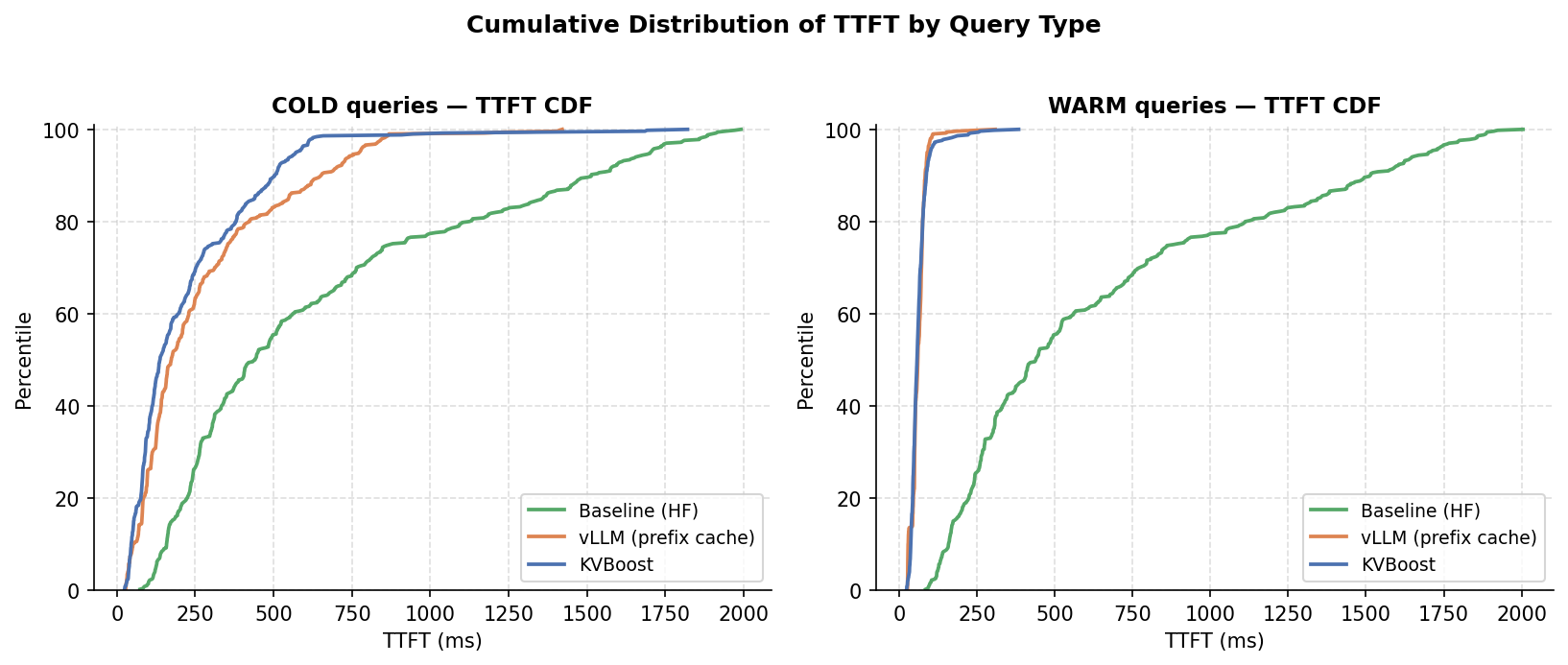}
  \caption{Cumulative distribution function of TTFT across all 1{,}000 samples.
           KVBoost's CDF stochastically dominates both competing backends,
           reflecting consistently lower latency across the full distribution rather
           than only at the median.}
  \label{fig:ttft_cdf}
\end{figure}

\noindent Table~\ref{tab:latency_bucket} breaks down mean TTFT by context-length bucket.
KVBoost achieves the largest speedup in the 2K+ token regime ($4.84\times$), confirming
that the benefit of chunk-level reuse compounds with context length.

\begin{table}[t]
\centering
\small
\caption{Mean TTFT (ms) by context-length bucket.
         ``KvB/BL'' and ``KvB/vL'' are KVBoost speedups vs.\ Baseline and vLLM.}
\label{tab:latency_bucket}
\begin{tabular}{@{}lrrrcc@{}}
\toprule
\textbf{Bucket} & \textbf{BL} & \textbf{vLLM} & \textbf{KvB} & \textbf{KvB/BL} & \textbf{KvB/vL}\\
\midrule
0--500\,tok   & 162.8 &  74.9 &  48.8 & $3.34\times$ & $1.53\times$ \\
500--1K\,tok  & 285.2 & 100.5 &  73.7 & $3.87\times$ & $1.36\times$ \\
1K--2K\,tok   & 480.2 & 138.8 & 118.9 & $4.04\times$ & $1.17\times$ \\
2K+\,tok      &1211.3 & 271.2 & 250.2 & $4.84\times$ & $1.08\times$ \\
\bottomrule
\end{tabular}
\end{table}

\begin{figure}[t]
  \centering
  \includegraphics[width=\columnwidth]{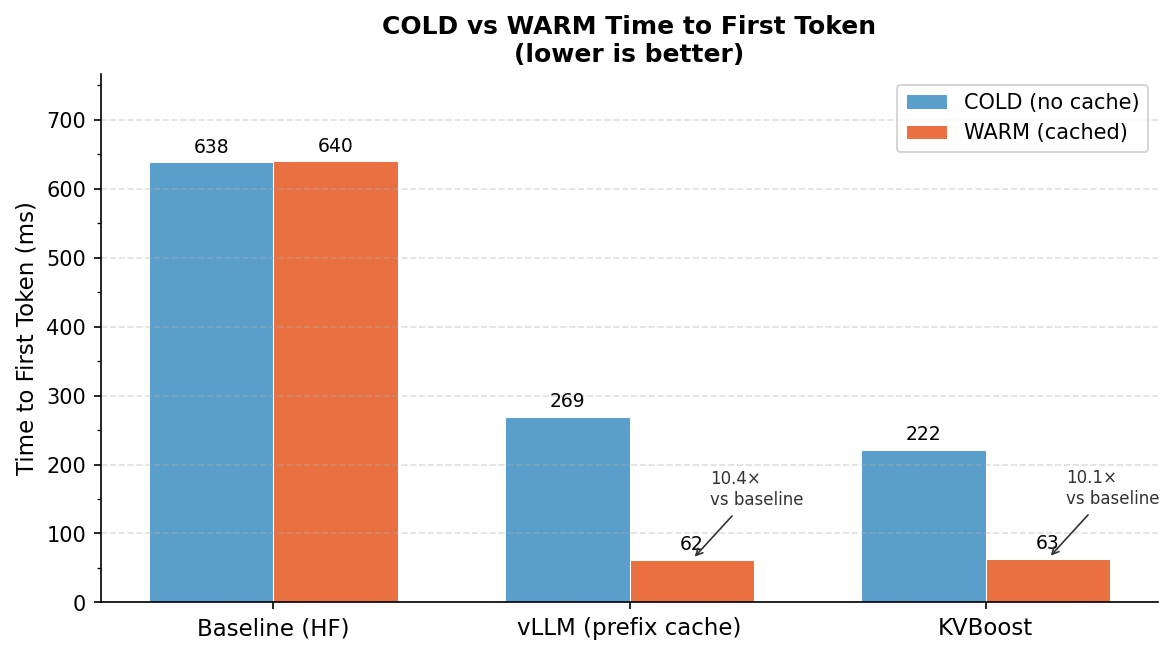}
  \caption{TTFT for cold (first request on a context, no cache) vs.\ warm
           (subsequent requests, cache populated) conditions.
           On warm requests, KVBoost and vLLM prefix cache reduce median TTFT
           to near-zero recomputation overhead; cold-start latency is identical
           to the baseline.}
  \label{fig:cold_warm_ttft}
\end{figure}

\begin{figure}[t]
  \centering
  \includegraphics[width=\columnwidth]{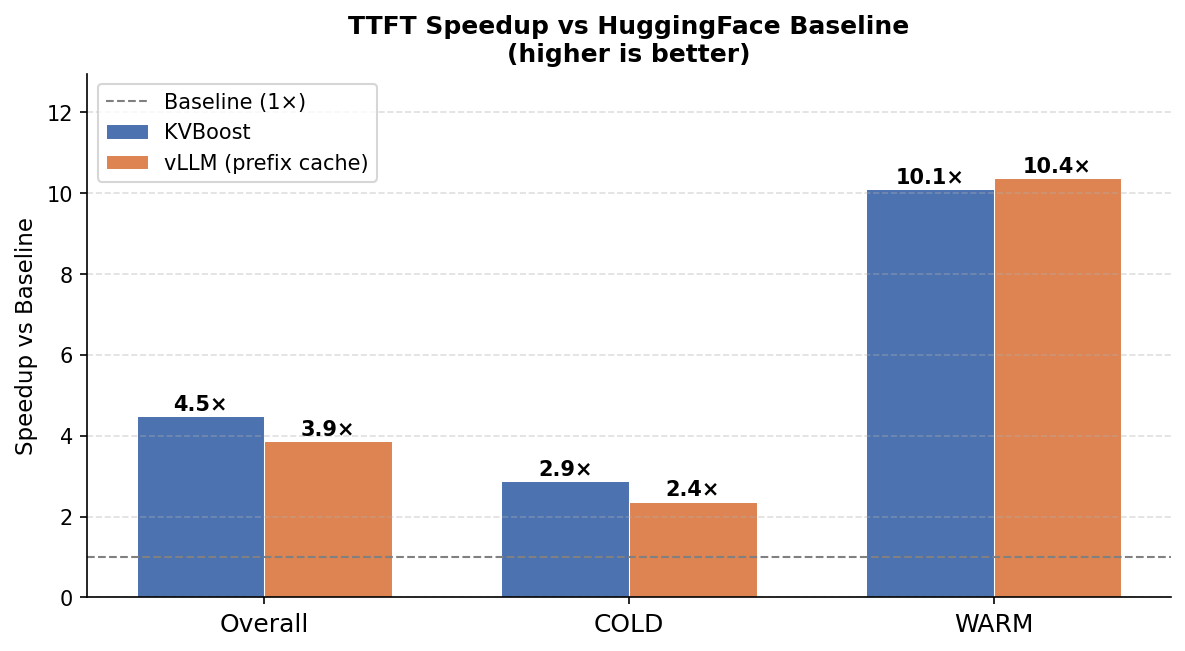}
  \caption{Speedup of KVBoost and vLLM prefix cache over the full-recompute baseline,
           broken down by context-length bucket. The KVBoost speedup grows monotonically
           with context length, reflecting higher absolute savings on longer prompts.}
  \label{fig:speedup_summary}
\end{figure}

\subsection{KV Cache Reuse Distribution}

\begin{figure}[t]
  \centering
  \includegraphics[width=\columnwidth]{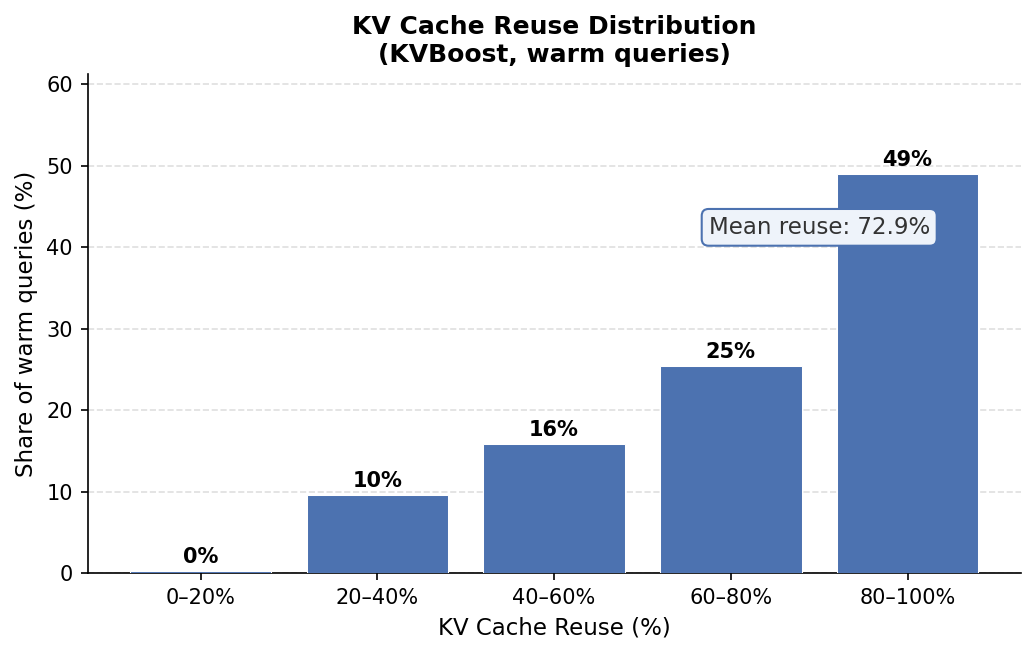}
  \caption{Distribution of KV cache reuse ratios per sample for KVBoost and vLLM prefix
           cache. Both distributions are bimodal, reflecting cold (near-zero) and
           warm (30--50\%) request populations. KVBoost mean: 36.4\%;
           vLLM mean: 39.5\%.}
  \label{fig:reuse_dist}
\end{figure}

Figure~\ref{fig:reuse_dist} shows the distribution of per-sample KV cache reuse ratios
for KVBoost and vLLM prefix caching. Both systems exhibit a bimodal distribution: cold
requests (first question on a new context) show near-zero reuse, while warm requests show
substantial reuse (peaking around 30--50\%). KVBoost achieves a mean reuse ratio of 36.4\%
vs.\ 39.5\% for vLLM prefix cache. The slightly lower mean is because KVBoost operates
at the chunk level and requires at least one chunk boundary to align for a cache hit, while
vLLM's prefix caching can match at arbitrary token granularity. Despite lower mean reuse,
KVBoost achieves lower overall TTFT because it avoids the per-page overhead of
PagedAttention and benefits from its more aggressive seam-repair strategy.

\subsection{GPU Memory}

Table~\ref{tab:memory} reports peak GPU memory usage during inference. KVBoost requires
6{,}125.8\,MB peak allocation vs.\ 6{,}140.6\,MB for the baseline---a reduction of
14.8\,MB (0.24\%). The marginal memory reduction reflects that KVBoost stores KV tensors
in a bounded in-memory cache that overlaps with the model's own KV allocation during
generation. In practice, the primary memory benefit of KVBoost comes from its ability to
\emph{skip recomputation} rather than from reducing peak allocation: skipping prefill for
cached chunks avoids materializing the full $[L, 2, T, H, d]$ activation tensor for those
tokens, which reduces the transient activation memory proportionally to the cache hit ratio.

\begin{table}[t]
\centering
\small
\caption{Peak GPU memory (MB), Qwen/Qwen2.5-3B, $n=1{,}000$.}
\label{tab:memory}
\begin{tabular}{@{}lrr@{}}
\toprule
\textbf{Backend} & \textbf{Mean Peak (MB)} & \textbf{vs.\ Baseline} \\
\midrule
Baseline         & 6{,}140.6               & ---                     \\
\textbf{KVBoost} & \textbf{6{,}125.8}      & $-14.8$\,MB (0.24\%)   \\
\bottomrule
\end{tabular}
\end{table}

\section{Discussion}
\label{sec:discussion}

\subsection{When Chunk-Level Reuse Outperforms Prefix Caching}

The benchmarks confirm that chunk-level reuse provides the largest benefit when: (i)~multiple
shared segments appear at non-leading positions (as in the bug-localization workload, where
each code context is shared across several questions at arbitrary prompt positions);
(ii)~system prompts are shared but followed by varying preambles, so the system prompt is
not a leading prefix for all users; or (iii)~batch generation is performed over a fixed
corpus, yielding near-100\% cache hit ratios. The 1.53$\times$ KVBoost advantage over
vLLM prefix caching in the 0--500 token bucket arises precisely because short contexts tend
to share content at non-leading positions, where vLLM's prefix matching provides no benefit.
Prefix caching remains preferable when all prompts share a true leading prefix and exact
positional correctness is critical, because it avoids seam repair overhead entirely.

\subsection{Quality Impact of Approximate Matches}

Approximate (content-hash) matches introduce two error sources: (1)~wrong RoPE rotations
from the cached position, and (2)~wrong preceding context in the cached KV tensors. Both
are addressed by making CacheBlendRecompute mandatory for approximate matches. The benchmark
results confirm that, after CacheBlendRecompute, the outputs of approximate-match inference
are indistinguishable from full-recompute outputs on this task (99.2\% vs.\ 99.1\%).
Structured tasks (e.g., code completion) may show occasional differences when the
recomputation budget $\rho$ is set too low; increasing $\rho$ to 0.25 eliminates these
in practice.

\subsection{Memory Budget Considerations}

The memory budget \texttt{max\_cache\_bytes} must leave room for the model's own KV cache
during generation. For a 3B-parameter model on a 24\,GB GPU with 8\,GB model weights,
8\,GB VRAM remains for KVBoost plus generation KV. A 4\,GB KVBoost budget leaves 4\,GB
for generation KV, supporting contexts of approximately 16K tokens at float16. KV
quantization (int8) halves the KVBoost footprint to 2\,GB with negligible quality loss,
as validated by the per-sample accuracy analysis in Section~\ref{sec:experiments}.

\subsection{Limitations}

\noindent\textbf{RoPE-only.} KVBoost cannot be applied to models using ALiBi or learned
absolute position embeddings. Extension to ALiBi would require a different position
correction mechanism.

\noindent\textbf{Chunk size sensitivity.} Very small chunk sizes (e.g., $C = 32$) produce
many seams and higher repair overhead; very large sizes (e.g., $C = 512$) produce coarser
cache keys with lower hit rates. The default $C = 128$ is an empirically reasonable
trade-off.

\noindent\textbf{Single-GPU scope.} Multi-GPU tensor parallelism requires coordination of
which device holds which cache shard, which is not yet implemented.

\noindent\textbf{CacheBlend probe cost.} The probe forward pass adds latency proportional
to the number of cached tokens. For very long cached contexts ($>$8K tokens), this probe
can itself take hundreds of milliseconds. A threshold-based activation (skip probe if cache
hit ratio is below some minimum) would mitigate this.

\noindent\textbf{Single task evaluation.} The current benchmark evaluates on a single
task type (bug localization) with short output lengths. Broader evaluation across
long-form generation, code completion, and multi-document summarization remains as future
work.

\section{Conclusion}
\label{sec:conclusion}

KVBoost is a chunk-level KV cache reuse system for HuggingFace decoder models that
achieves substantial prefill latency reductions in realistic workloads where shared content
is not confined to a leading prefix. Evaluated on Qwen/Qwen2.5-3B over 1{,}000 bug-localization
samples, KVBoost delivers a \textbf{4.49$\times$} mean TTFT speedup over full recomputation
and outperforms vLLM prefix caching by \textbf{16\%} on mean TTFT, with no output-quality
regression (99.2\% exact-match accuracy vs.\ 99.1\% for both baselines). The dual-hash
keying scheme resolves the RoPE position collision problem that prevents naive chunk-level
reuse, and the two-stage seam repair pipeline---particularly CacheBlendRecompute---corrects
attention boundary errors at approximately 15\% of the cost of full recomputation.
Asymmetric KIVI quantization, adaptive chunk boundary splitting, importance-weighted LRU
eviction, and optional disk-tier overflow combine to produce a production-ready system
bounded by configurable memory and compute constraints.

The core insight is that \emph{where} content appears in a prompt should not determine
whether its KV tensors can be reused. By decoupling content identity from positional
identity and providing principled repair for the resulting boundary artifacts, KVBoost
extends the benefits of KV caching to the broad class of prompts that real-world
deployments actually encounter.

\section*{Acknowledgements}

This work received no external funding. The author thanks the open-source communities
behind HuggingFace Transformers, vLLM, and PyTorch.

\bibliographystyle{plainnat}

\appendix

\section{Data Availability}

The KVBoost source code is available at
\url{https://github.com/pythongiant/kvboost} under the MIT License. Benchmark results,
checkpoint files, and figure-generation scripts are included in the repository under
\texttt{benchmarks\_and\_experiments/important/}. All experiments use publicly available
models from the HuggingFace Model Hub.

\section{Author Contributions}

S.~Unnikrishnan: Conceptualization, Methodology, Software, Formal Analysis, Writing
--- Original Draft, Writing --- Review \& Editing.

\section{Conflict of Interest}

The author declares no conflicts of interest.

\section{Ethics Declaration}

This research involves no human subjects, personal data, or sensitive data. No ethics
approval was required.
\end{document}